\pdfoutput=1
\documentclass[11pt]{article}

\usepackage[final]{ACL2023}

\usepackage{times}
\usepackage{latexsym}
\usepackage{graphicx}
\usepackage{float}
\usepackage{tablefootnote}  % For footnotes inside tables
\usepackage{caption}        % Better caption control
\usepackage{graphicx} 
\usepackage{tikz} 
\usepackage{hyperref}

\usepackage[T1]{fontenc}
\usepackage[utf8]{inputenc}

\usepackage{microtype}

\usepackage{inconsolata}

\title{L3Cube-IndicHeadline-ID: A Dataset for Headline Identification and Semantic Evaluation in Low-Resource Indian Languages}
\author{
  Nishant Tanksale\textsuperscript{1,3} \quad
  Tanmay Kokate\textsuperscript{1,3} \quad
  Darshan Gohad\textsuperscript{1,3} \\
  \textbf{Sarvadnyaa Barate\textsuperscript{1,3}} \quad
  \textbf{Raviraj Joshi\textsuperscript{2,3}}\thanks{Correspondence: ravirajoshi@gmail.com} \\
  \textsuperscript{1}Department of Information Technology, PICT, Pune \\
  \textsuperscript{2}Indian Institute of Technology Madras, Chennai \\
  \textsuperscript{3}L3Cube Labs, Pune \\
}

\begin{document}
\maketitle
\begin{abstract}

Semantic evaluation in low-resource languages remains a major challenge in NLP. While sentence transformers have shown strong performance in high-resource settings, their effectiveness in Indic languages is underexplored due to a lack of high-quality benchmarks. To bridge this gap, we introduce L3Cube-IndicHeadline-ID, a curated headline identification dataset spanning ten low-resource Indic languages: Marathi, Hindi, Tamil, Gujarati, Odia, Kannada, Malayalam, Punjabi, Telugu, Bengali and English. Each language includes 20,000 news articles paired with four headline variants: the original, a semantically similar version, a lexically similar version, and an unrelated one, designed to test fine-grained semantic understanding. The task requires selecting the correct headline from the options using article-headline similarity. We benchmark several sentence transformers, including multilingual and language-specific models, using cosine similarity. Results show that multilingual models consistently perform well, while language-specific models vary in effectiveness. Given the rising use of similarity models in Retrieval-Augmented Generation (RAG) pipelines, this dataset also serves as a valuable resource for evaluating and improving semantic understanding in such applications. Additionally, the dataset can be repurposed for multiple-choice question answering, headline classification, or other task-specific evaluations of LLMs, making it a versatile benchmark for Indic NLP. The dataset is shared publicly at \url{https://github.com/l3cube-pune/indic-nlp}.
\end{abstract}

\section{Introduction}

Natural Language Processing (NLP) faces persistent challenges in addressing low-resource languages, primarily due to a lack of standardized datasets and methodologies. Despite their cultural and linguistic significance, many Indian languages remain underrepresented in research, creating barriers to developing effective solutions~\cite{dongare2024indianlanguages,magueresse2020lowresource}.

Social media platforms, in particular, underscore the pressing need for tailored NLP solutions. Text-driven interactions in regional languages dominate these platforms, underscoring the need for semantic understanding technologies that can process and analyse the growing volume of regional language data. Concurrently, the emergence of Retrieval-Augmented Generation (RAG) methods has driven the need for high-quality sentence embeddings to enhance information retrieval and generation tasks \citep{panchal-shah-2024-nlp}

Sentence transformers, with their ability to generate semantically meaningful embeddings, provide a promising avenue for advancing NLP in low-resource contexts. By incorporating pooling layers over the outputs of BERT-based models, these transformers produce fixed-size embeddings that are both efficient and effective for downstream tasks~\citep{reimers2019sentencebert}.

Recent years have witnessed the development of BERT-based models for Indic languages, reflecting the growing need for NLP tools catering to non-English-speaking populations~\cite{deode2023indicSBERT}. However, a significant gap persists in creating robust datasets for assessing these models in multilingual and low-resource contexts.

To address this gap, we introduce the L3Cube-IndicHeadline-ID\footnote{\href{https://huggingface.co/datasets/l3cube-pune/IndicHeadline-ID}{l3cube-pune/IndicHeadline-ID}}: a novel benchmark comprising 20,000 news articles for each of ten major Indic languages~\cite{mirashi2024indicnews}. Each article is paired with four titles: the original, a semantically similar title, a lexically similar title, and an unrelated title. This dataset evaluates models based on their ability to identify the original title using cosine similarity, offering a scalable and nuanced evaluation approach for sentence-level semantic understanding in Indic languages. We particularly focus on the semantic understanding aspect due to the growing importance of similarity models in recent Retrieval-Augmented Generation (RAG) applications. Beyond semantic similarity evaluation, the dataset can also be used for headline identification using classification models or framed as a multiple-choice question answering task, making it versatile for various natural language understanding applications.

\section{Related Work}

Semantic evaluation in multilingual and low-resource languages has gained significant attention in recent years. Foundational efforts such as the Semantic Textual Similarity (STS) task and the Semantic Textual Relatedness (STR) dataset have provided valuable sentence pair annotations across multiple languages~\cite{ousidhoum2024semrel,conneau-etal-2018-xnli}. However, these resources predominantly emphasize high-resource languages and rely on extensive manual annotations, limiting their applicability and scalability in low-resource contexts.

For Indic languages, the IndicNLP Suite introduced by~\cite{kakwani2020indicnlpsuite} has made notable strides by offering pre-trained language models and evaluation benchmarks. However, its headline prediction dataset is not publicly available, creating a gap in publicly available resources for semantic evaluation in Indian languages. Similarly, multilingual datasets like MASSIVE have extended coverage to 51 languages but are primarily designed for token-level tasks such as intent recognition and slot filling, leaving sentence-level evaluation in Indic languages largely unaddressed~\cite{fitzgerald2023massive}.

Efforts such as IndicSentEval by~\cite{aravapalli2024indicsenteval} further advance the field by evaluating the linguistic properties encoded by multilingual transformer models. Their dataset and probing tasks focus on surface, syntactic, and semantic features across six Indic languages, shedding light on the strengths of both universal and Indic-specific models. However, this work emphasizes probing linguistic representations rather than sentence-level semantic alignment.

Efforts such as IndicMT Eval~\citep{SaiB2023IndicMTEval} have explored meta-evaluation frameworks for machine translation metrics, addressing the limitations of standard MT evaluation methods for Indian languages. The dataset captures linguistic and cultural nuances across multiple Indic languages, providing a benchmark for translation quality assessment.

Our work complements and builds upon these initiatives by introducing the first publicly available large-scale dataset specifically designed for headline identification across ten Indic languages~\cite{mirashi2024indicnews}. By systematically generating diverse candidate headlines, we address limitations associated with manual annotations. Our framework provides a robust, scalable mechanism for evaluating sentence transformers, offering a unique contribution to sentence-level semantic evaluation in low-resource Indic languages.

\begin{figure*}
    \centering
    \includegraphics[scale=0.6]{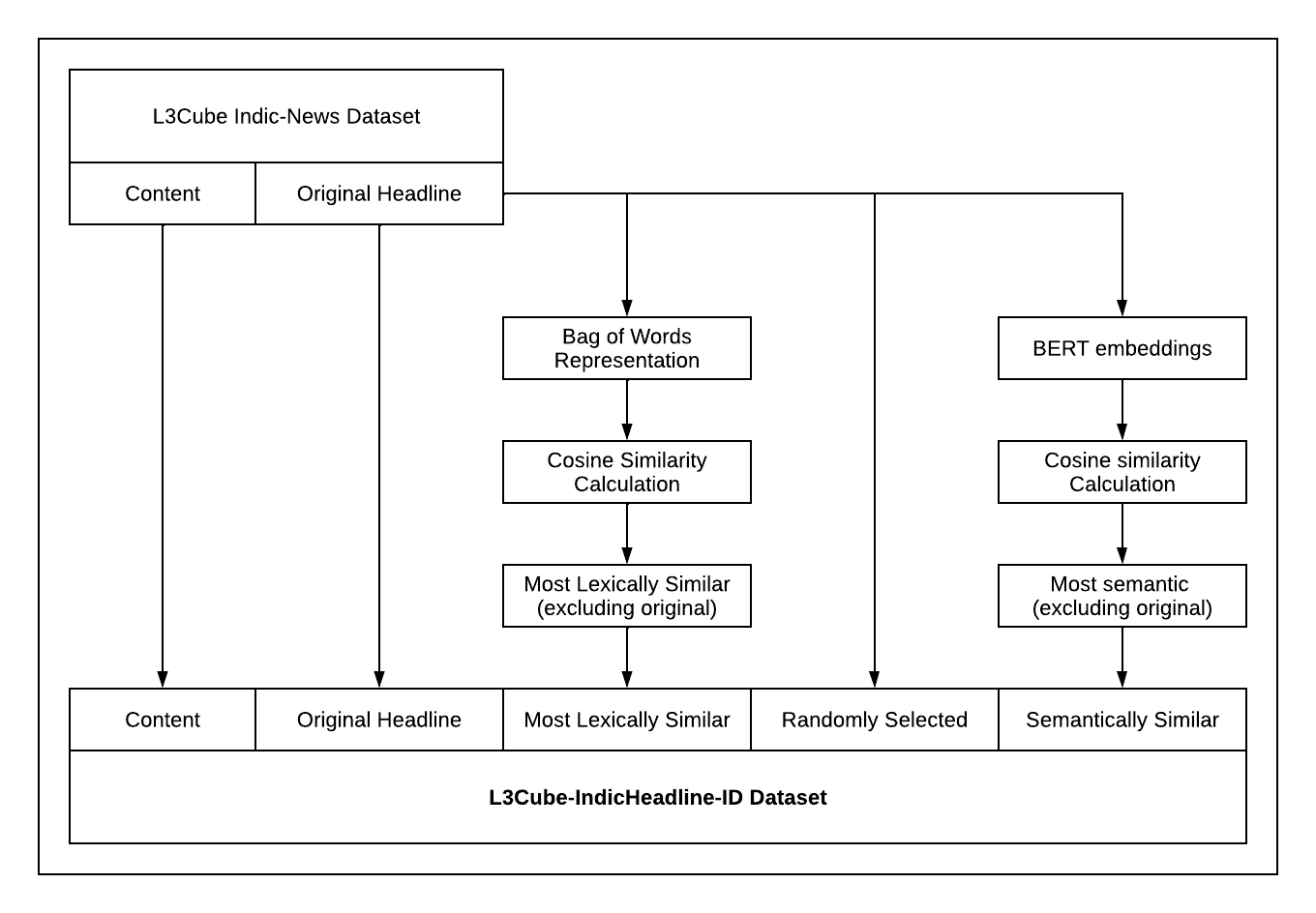}
    \caption{Methodology for creating the dataset with candidate titles}
    \label{fig:dataset_generation}
\end{figure*}

\begin{table*}[ht]
    \centering
    \resizebox{\textwidth}{!}{
    \begin{tabular}{|p{4cm}|p{4cm}|p{4cm}|p{4cm}|}
        \hline
        \textbf{Original Title / Headline} & \textbf{Semantic Title} & \textbf{Lexical Title} & \textbf{Random Title} \\
        \hline
        Chhattisgarh Polls: Congress Releases First List, Fields CM Bhupesh Baghel From Patan & 
        Ahead Of Phase-1 Polling In Chhattisgarh, CM Baghel Faces The Heat For Mahadev App Case. Top Points & 
        Chhattisgarh CM Bhupesh Baghel Resigns After Congress Party's Shock Defeat & 
        Chandrayaan-3 Launches Today: 10 Interesting Facts About ISRO's Third Moon Mission \\
        \hline
        Hyundai Ioniq 5 Electric SUV With 631 Km Range — First Look Review & 
        Hyundai Ioniq 5 First Drive Review: Lots Of Tech, Space and Style — Know If This EV Is Value-For-Money & 
        BMW iX Electric SUV First Look Review — Know About Design And Interiors & 
        North India Likely To Usher In New Year With More Chill and Rain, No Respite In Sight From Fog Till Saturday \\
        \hline
    \end{tabular}
    }
    \caption{Examples of English headlines with semantic, lexical, and random negatives.}
    \label{tab:english-examples}
\end{table*}

\section{Methodology}

The dataset used in this study is based on L3Cube-IndicNews~\cite{mirashi2024indicnews} and includes 20,000 samples for each of ten Indic languages: Marathi, Hindi, Tamil, Gujarati, Odia, Kannada, Malayalam, Punjabi, Telugu, and Bengali. Each sample comprises a news article paired with its original headline.

To construct IndicHeadline-ID, a headline identification dataset, three additional candidate headlines were selected for each article, forming a set of four options: the original headline, a semantically similar headline, a lexically similar headline, and a random headline. The original serves as the ground truth.
This setup enables a nuanced evaluation of sentence transformers, especially for retrieval-augmented generation (RAG) tasks, by testing their ability to capture semantic meaning, distinguish lexical overlap, and reject irrelevant content. The candidate titles are selected as follows:

\begin{itemize}
    \item \textbf{Original Title}: The true headline of the news article, directly sourced from the IndicNews dataset. It serves as the ground truth and is used to assess the model's ability to accurately align embeddings with the article's context.
    \item \textbf{Semantically Similar Title}: A different title that expresses the same core meaning as the original but might use different words or phrasing. These titles are selected using language-specific sentence embedding models developed by L3Cube Labs \cite{deode2023indicSBERT}. Cosine similarity is computed between the embedding of the original title and all other titles in the dataset. The most semantically similar title is selected, excluding the original itself, for each instance. This candidate is included to challenge the model by introducing a plausible yet incorrect option, making the selection task more difficult and testing the model’s ability to distinguish between closely related semantic content.
    \item \textbf{Lexically Similar Title}: A title that shares significant word overlap with the original headline but differs in meaning. These titles are identified by computing word frequency-based vector representations for all titles and retrieving the one with the highest lexical similarity to the original. This comparison captures similarity at the surface level of word usage, without incorporating deeper semantic relationships.
    \item \textbf{Random Title}: An unrelated headline arbitrarily chosen from the dataset. These distractor titles are included to evaluate the model's robustness in rejecting irrelevant content and simulating realistic conditions with unrelated information. 
\end{itemize}

The examples for English are provided in Table~\ref{tab:english-examples}.

% \begin{table*}[ht]
%     \centering
%     \begin{tabular}{|p{4cm}|p{4cm}|p{4cm}|p{4cm}|}
%         \hline
%         \textbf{Title} & \textbf{Semantic Title} & \textbf{Lexical Title} & \textbf{Random Title} \\
%         \hline
%         Chhattisgarh Polls: Congress Releases First List, Fields CM Bhupesh Baghel From Patan & 
%         Ahead Of Phase-1 Polling In Chhattisgarh, CM Baghel Faces The Heat For Mahadev App Case. Top Points & 
%         Chhattisgarh CM Bhupesh Baghel Resigns After Congress Party's Shock Defeat & 
%         Chandrayaan-3 Launches Today: 10 Interesting Facts About ISRO's Third Moon Mission \\
%         \hline
%         Hyundai Ioniq 5 Electric SUV With 631 Km Range — First Look Review & 
%         Hyundai Ioniq 5 First Drive Review: Lots Of Tech, Space and Style — Know If This EV Is Value-For-Money & 
%         BMW iX Electric SUV First Look Review — Know About Design And Interiors & 
%         North India Likely To Usher In New Year With More Chill and Rain, No Respite In Sight From Fog Till Saturday \\
%         \hline
%     \end{tabular}
%     \caption{Examples of English headlines with semantic, lexical, and random negatives.}
%     \label{tab:english-examples}
% \end{table*}

\begin{table*}[ht]
    \centering
    \begin{tikzpicture}[remember picture,overlay]
        \node[rotate=90, anchor=north west] at (current page.north east) [xshift=-1cm, yshift=-6cm] {\footnotesize x: language-specific model};
    \end{tikzpicture}

    \begin{tabular}{|l|c|c|c|c|c|c|}
        \hline
        \multicolumn{5}{|c|}{\textbf{Multilingual Models}} & \multicolumn{2}{c|}{\textbf{Language-Specific Models}} \\
        \hline
        & \shortstack{\textbf{indic-} \\ \textbf{sentence-} \\ \textbf{bert-nli}}
        & \shortstack{\textbf{indic-} \\ \textbf{sentence-} \\ \textbf{similarity-sbert}}
        & \shortstack{\textbf{multilingual-} \\ \textbf{e5-base}}
        & \shortstack{\textbf{muril base} \\ \textbf{cased}}
        & \shortstack{\textbf{x-bert} \\ \textbf{}}
        & \shortstack{\textbf{x-sentence-} \\ \textbf{similarity-} \\ \textbf{sbert}} \\
        \hline
        \textbf{Marathi} & 0.5756 & 0.5561 & 0.5747 & 0.5831  & \textbf{0.5948} & 0.5579 \\
        \hline
        \textbf{Hindi} & 0.8428 & 0.8514 & \textbf{0.9089} & 0.7449  & 0.7794 & 0.8625 \\
        \hline
        \textbf{Tamil} & 0.8415 & 0.8481 & 0.8815 & 0.7645 &  0.7543  & \textbf{0.8574} \\
        \hline
        \textbf{Gujarati} & 0.8096 & 0.8082 & \textbf{0.8390} & 0.7051 &  0.7392 & 0.8162 \\
        \hline
        \textbf{Odia} & 0.8210 & 0.7891 & \textbf{0.8468} & 0.6136 &  0.6732  & 0.8039 \\
        \hline
        \textbf{Kannada} & 0.8650 & 0.8730 & 0.8912 & 0.7953 &  0.7935  & \textbf{0.8918} \\
        \hline
        \textbf{Malayalam} & 0.8092 & 0.8102 & 0.8155 & 0.6905 &  0.6891 & \textbf{0.8239} \\
        \hline
        \textbf{Punjabi} & 0.9712 & 0.9609 & 0.9715 & 0.9257  & 0.9193 & \textbf{0.9728} \\
        \hline
        \textbf{Telugu} & 0.8075 & 0.8090 & \textbf{0.8400} & 0.7129 &  0.7270  & 0.6395 \\
        \hline
        \textbf{Bengali} & 0.8009 & 0.8123 & \textbf{0.8385} & 0.7115 &  0.6897 & 0.8305 \\
        \hline
        \textbf{English} & 0.6554 & 0.6312 & \textbf{0.7681} & -- & -- & -- \\
        \hline
    \end{tabular}

    \caption{Performance of Multilingual and Language-Specific Models on Different Languages (x:language).}
    \label{tab:multilingual_language_models}
\end{table*}

During evaluation, each news article and its candidate headlines were encoded into embeddings using sentence transformer models ~\citep{deode2023indicSBERT}. Cosine similarity scores were calculated between the article embedding and each candidate headline embedding, and the candidate with the highest score was identified as the predicted match ~\citep{lin2017structured,mueller2016siamese}. This dataset-based evaluation eliminates the need for manual similarity annotations, which are challenging to scale for low-resource languages.

The algorithmic generation of semantic and lexical candidates ensures consistency across languages and scalability. By including all headline types, the dataset provides a detailed assessment of model strengths and weaknesses, which cannot be achieved with single-score metrics like BLEU or METEOR ~\citep{denkowski-lavie-2014-meteor}. This approach offers valuable insights into sentence transformers' performance on low-resource Indic languages and paves the way for further advancements in the domain.

\section{Models}
We benchmarked a set of multilingual and Indic-specific sentence transformer models to evaluate semantic understanding in low-resource Indic languages. Multilingual models like  \textit{multilingual-e5-base} ~\citep{wang2022e5} and \textit{google-muril-base-cased} ~\citep{khanuja2021muril} were chosen for their strong cross-lingual transfer capabilities and proven performance on semantic similarity tasks. Indic-specific models such as \textit{IndicSBERT} ~\citep{deode2023indicSBERT}and \textit{language specific BERT} ~\citep{joshi2022l3cubehind} were included as they are pre-trained or fine-tuned on Indian language corpora, making them more attuned to the linguistic characteristics of the region. This diverse selection allowed us to compare general-purpose multilingual models with those explicitly designed for Indic languages. The goal was to understand how well each model captures sentence-level semantics in scenarios with limited language resources.

\section{Results and Discussion}

Among the multilingual models, indic-sentence-bert-nli and indic-sentence-similarity-sbert demonstrated consistent performance across most languages (see Table \ref{tab:multilingual_language_models}). However, multilingual-e5-base outperformed other models for Hindi and Tamil, achieving cosine similarity scores of 0.9089 and 0.8815, respectively, and also showed strong results for English.

Language-specific models generally achieved higher scores than multilingual counterparts for certain languages. For instance, the Kannada-specific model obtained the highest score of 0.8914. In contrast, Marathi and Bengali exhibited comparatively lower performance across all models, reflecting the challenges of semantic evaluation in these contexts.

Overall, the results highlight the strengths of language-specific models in capturing nuanced semantics within individual languages, while multilingual models provide a robust baseline for cross-lingual evaluation.

\section{Limitations}

While the Indic Headline Identification Dataset ensures linguistic consistency by sourcing news articles in formal dialects, it does not account for informal or dialectal variations. This limitation may impact the generalizability of findings to less formal contexts, such as social media or colloquial speech.

Additionally, the dataset’s reliance on algorithmically selected candidates, while scalable, may not capture the full spectrum of semantic and lexical diversity observed in real-world scenarios. Future work could explore incorporating manually annotated datasets or augmenting the framework with synthetic data to address these limitations.

\section{Conclusion}
In this work, we present L3Cube-IndicHeadline-ID, a comprehensive dataset and evaluation framework for semantic similarity across ten Indic languages and English, filling a crucial gap in the landscape of low-resource NLP. Through extensive benchmarking of state-of-the-art sentence transformers, we offer key insights into the strengths and limitations of both multilingual and language-specific models.

Our results demonstrate the strong performance of models like multilingual-e5-base, while also underscoring the benefits of targeted fine-tuning for individual languages. These findings emphasize the importance of balancing generalization with specialization in multilingual NLP. Moreover, the persistent challenges in low-resource settings call for future research into more adaptable multilingual architectures and equitable pre-training paradigms that better serve diverse linguistic communities.

By releasing L3Cube-IndicHeadline-ID and associated benchmarks, we provide an essential resource to catalyze further research in Indic semantic similarity. We hope this work will drive the development of more inclusive and representative language technologies, empowering NLP applications across the linguistically rich and diverse Indian subcontinent.

\section*{Acknowledgment}
This work was carried out under the mentorship of L3Cube Labs, Pune. We would like to express our gratitude towards our mentor for his continuous support and encouragement. This work is a part of the L3Cube-IndicNLP Project.

\bibliographystyle{acl_natbib}

\bibliography{main}

\end{document}